\pdfoutput=1

\documentclass[11pt]{article}

\usepackage{ACL2023}

\usepackage{times}
\usepackage{latexsym}
\usepackage{diagbox}
\usepackage{multirow}
\usepackage{ulem}
\usepackage{pdfrender}
\usepackage{url}
\usepackage{enumitem}
\usepackage{cancel}
\usepackage{longtable}
\usepackage{times}
\usepackage{latexsym}
\usepackage{enumitem}
\usepackage[T1]{fontenc}
\usepackage[utf8]{inputenc}
\usepackage{microtype}
\usepackage{inconsolata}
\usepackage{times}
\usepackage{latexsym}
\usepackage{comment}
\usepackage{adjustbox}
\usepackage{booktabs}
\usepackage{subcaption}
\usepackage[english]{babel}
\usepackage[utf8]{inputenc}
\usepackage{amsmath}
\usepackage{graphicx}
\usepackage{epsfig}

\usepackage[T1]{fontenc}

\usepackage[utf8]{inputenc}

\usepackage{microtype}

\usepackage{inconsolata}

\title{Advancing Post-OCR Correction: \\A Comparative Study of Synthetic Data}

\author{Shuhao Guan, Derek Greene \\
  Insight Centre for Data Analytics, Dublin \\
  School of Computer Science, University College Dublin, Ireland \\
  \texttt{shuhao.guan@ucdconnect.ie, derek.greene@ucd.ie} 
  }

\begin{document}
\maketitle
\begin{abstract}
This paper explores the application of synthetic data in the post-OCR domain on multiple fronts by conducting experiments to assess the impact of data volume, augmentation, and synthetic data generation methods on model performance. Furthermore, we introduce a novel algorithm that leverages computer vision feature detection algorithms to calculate glyph similarity for constructing post-OCR synthetic data. Through experiments conducted across a variety of languages, including several low-resource ones, we demonstrate that models like ByT5 can significantly reduce Character Error Rates (CER) without the need for manually annotated data, and our proposed synthetic data generation method shows advantages over traditional methods, particularly in low-resource languages\footnote{Code and data are available at \url{https://github.com/NikoGuan/P_OCR}}.
\end{abstract}

\section{Introduction}
\label{sec:intro}

Digital libraries, like the Internet Archive, offer a vast collection of historical and culturally important books in image formats, including works written in low-resource and endangered languages. However, their image-only format limits content accessibility, hindering the use of these essential resources.
%
%
Therefore, Optical Character Recognition (OCR) technologies are evidently useful in this context. However, OCR outputs frequently contain errors, particularly when working with texts featuring complex styles, archaic fonts, or unconventional layouts. These errors may include character recognition mistakes, formatting issues, and hyphenation problems, which are particularly prominent when dealing with low-resource languages \cite{ignat2022ocr}. Poor quality OCR can reduce the usefulness of these digital texts, adversely affecting downstream tasks \citep{linhares2019impact,koudoro2021spatial}.

Post-OCR correction is crucial for multiple reasons, including cultural heritage preservation \cite{jarlbrink2017cultural}, expanding the accessibility of knowledge and information \cite{bazzo2020assessing}, and supporting further downstream tasks in cultural analytics \cite{stubbs1996text}. Additionally, accurate historical text data is extremely important for training large language models (LLMs) \cite{bubeck2023sparks}, which require high-quality data to improve their ability to handle complex queries, especially those involving history and culture.

Crowdsourcing has been the primary approach to acquire post-OCR training data in this domain \cite{clematide2016crowdsourcing,richter2018low,maheshwari2022benchmark}. While this can provide highly accurate training data, the process is often time consuming and expensive. With the advent of Transformer architecture and attention mechanisms \cite{vaswani2017attention}, deep learning models have emerged as the standard approach for post-OCR tasks. These models require large amounts of data for training. Thus, synthetic data has been increasingly adopted in this context. \cite{d2017generating,davydkin2023data,jasonarson2023generating}. However, most existing literature on generating synthetic data relies on additional existing data for generation, and no comprehensive comparison has been made between different methods to understand how the synthetic data generated in various ways affects post-OCR performance.

To address these issues, this paper explores the impact of data volume and data augmentation methods on the performance of post-OCR models. We examine several common methods for creating synthetic data in the post-OCR domain and propose a novel method based on feature detection algorithms from computer vision to calculate glyph similarity for synthetic data construction. We conduct experiments on eight languages, including several low-resource languages, achieving significant CER reductions ranging from 12.41\% to 48.18\%.


\section{Related Work}
\label{sec:related}
Popular OCR systems include the Google Vision API OCR system \cite{fujii2017sequence} and the Tesseract OCR engine \cite{smith2007overview}. 
\citet{jatowt2019deep} performed statistical analysis on the types of errors commonly produced by OCR systems.

Post-OCR correction, while often overlooked, is an important NLP task. Lexical approaches to post-correction concentrate on character and word level inaccuracies, primarily employing dictionaries and heuristic rules. \citet{bassil2012ocr} exploited Google's search suggestions for context-based corrections, circumventing the need for exhaustive dictionaries. Strategies specific to certain domains, such as those proposed by \citet{furrer2011reducing}, \citet{estrella2014ocr}, and \citet{kettunen2016keep}, highlight the necessity of tailored dictionaries for texts that possess unique features, like historical typefaces. \citet{wemhoener2013creating} focused on aligning and merging outputs from various scans, including those from different editions, to rectify errors. 
Recent studies have framed post-OCR tasks as Seq2Seq tasks, with researchers applying both  Statistical Machine Translation (SMT) and Neural Machine Translation (NMT) models \citep{amrhein2018supervised}. \citet{nguyen2020neural} and \citet{soper2021bart} employed BERT \cite{devlin2018bert} and BART \cite{lewis2019bart} models, respectively. \citet{maheshwari2022benchmark} conducted a comparison between pre-trained models and traditional Seq2Seq models, with their findings indicating that pre-trained models like ByT5 \cite{xue2022byt5} outperform the conventional models. \citet{ramirez2022post} segmented documents into character n-grams, before aggregating their corrections into the final output via majority voting \cite{lam1997application}, essentially acting as an ensemble of sequence models.


Data is fundamental to the success of deep learning, as an increasing amount of research is directed towards leveraging data-driven strategies. These strategies aim to significantly improve model performance by optimizing data usage, rather than modifying the underlying model structure \cite{tarafdar2019using,mazumder2022dataperf}. These efforts often involve generating large quantities of synthetic data \cite{choi2023dmops}, involving data manipulation measures like filtering \cite{koehn2020findings}, data augmentation \cite{shorten2019survey,li2022data}, and noise injection \cite{izumi2003automatic}.
Such synthetic data has been widely used in various NLP tasks, such as grammatical error correction \cite{ingolfsdottir2023byte}, language identification \cite{ahmadi2023pali}, question answering \cite{puri2020training}, and named entity recognition \cite{liu2021mulda}.

For the task of text denoising, the primary method for constructing synthetic data is noise injection \cite{izumi2003automatic}. This technique involves inserting errors into clean text to generate training data pairs. \citet{d2017generating} added artificial OCR errors into sentences using a random process, while \citet{jasonarson2023generating} focused on the low-resource Icelandic language for post-OCR tasks. The authors extracted OCR errors from real digitised documents and inserting them into clean text in similar proportions. \citet{grundkiewicz2019neural} analyzed real data to create replacement sets for each word. \citet{davydkin2023data} adopted a different approach, developing a system to generate handwritten image data which were then processed with OCR. The resulting OCR outputs and the original texts were subsequently used to train a T5 model \cite{raffel2020exploring} for correction purposes. \citet{ignat2022ocr} synthesized text image datasets by manipulating parameters, like font spacing and image saturation, then compare the original text with the OCR output text, to evaluate OCR systems' performance on various low-resource languages, and they enhanced Machine Translation (MT) through backtranslation \cite{sennrich2015improving}.

Some studies have explored improving post-OCR text correction by analyzing the visual forms of characters, known as glyphs. \citet{chen2023enhancing} attempting to use the CharBERT model \cite{ma2020charbert} for post-OCR. Their method consists of two parallel CNN encoders and a transformer decoder, taking CharBERT and glyph embedding as inputs. \citet{amrhein2018supervised}'s experiment included NMT models and glyph embedding, but it did not enhance model performance. Other research has focused on the detection of homoglyphs, with \citet{ginsberg2018rapid} employing a grid method to assess the similarity of glyphs by counting the number of overlapping grids between characters. The similarity of glyphs is actually based on visual features. In the field of computer vision, feature detection and matching algorithms, such as SIFT \cite{ng2003sift}, ORB \cite{rublee2011orb}, and AKAZE \cite{alcantarilla2011fast} can extract feature points from an image and match them with those appearing in other images.

\section{Methods}
We now describe three common methods for generating synthetic data in the post-OCR domain, before introducing a new method based on \textit{glyph similarity} in Section \ref{sec:methods_4}. Each method makes use of a clean corpus \(A\). As a common preprocessing step, we segment \(A\) into multiple chunks by initially dividing sentences based on punctuation marks and then concatenating these sentences, ensuring that the total length of each chunk does not exceed a fixed limit of 230 characters, with the exception of Russian and Telugu, where the limits are 140 and 90 characters, respectively.

\label{sec:methods}

\subsection{Method $\textcircled{1}$ Random Injection}
\label{sec:methods_1}
The random injection method \cite{d2017generating} generates a synthetic OCR corpus $\hat{A}$ by randomly inserting errors into an existing clean corpus $A$.

First, we filter out infrequently occurring characters in $A$, as the corpus may inevitably contain some noise. The remaining characters are used for replacements and insertions.
%
%
Next, for each chunk from $A$, we randomly select a target error rate $p \in [0, 15]$, which controls the level of noise to be introduced. According to the analysis by \citet{jatowt2019deep}, the average rate of OCR errors involving substitution, insertion, and deletion is approximately 5:1:1. Therefore, in our implementation, each character has a probability of $\frac{5}{7} \times p$ of being replaced by a random character and a probability of $\frac{1}{7} \times p$ of being deleted. Additionally, any two characters have a probability of $\frac{1}{7} \times p$ of having a random character inserted between them.


\subsection{Method $\textcircled{2}$ Image Creation}
\label{sec:methods_2}

Following some OCR-related studies that add noise to text images \cite{jaderberg2014synthetic,krishna2018upcycle,boros2022assessing}, we also investigate the simulation of real-world correction scenarios by creating synthetic text images. Specifically, each chunk from corpus $A$ is used to create a separate image which is manipulated to add OCR-like noise. These images are then processed through a standard OCR system to obtain $\hat{A}$.

The complete procedure is as follows. Initially, for every text chunk, a random font from the set $F$, which matches the language of corpus $A$, is chosen.
Then, the text from each chunk is converted into an image. These images are subjected to random rotations of $\pm  5$ degrees, followed by the insertion of random noise pixels. Following this, the images undergo random dilation and erosion to simulate text variability, and their resolution are randomly reduced within a range of 0 to 50\%. At the end of this process, the chosen OCR system is applied for text recognition,  resulting in the output $\hat{A}$. 

\begin{figure}[!t]
  \centering
  \includegraphics[width=0.33\textwidth]{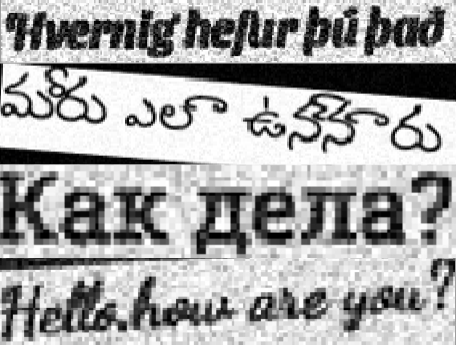}
  \caption{Examples of synthetic OCR images in various languages, generated using the process in Section \ref{sec:methods_2}.}
  \label{fig:ocr}
\end{figure}

Figure \ref{fig:ocr} shows sample synthesized images with noise generated using this process, for several different languages. Although synthetic images are used, this method simulates the OCR process under real-world conditions, 
the output text is derived from the OCR system, hence we consider it to be representative of authentic OCR text.The corpora $A$ and $\hat{A}$ can be used to construct test sets, training sets, or to extract distributions of OCR errors.

\begin{figure*}[!th]
  \centering
  \includegraphics[width=0.85\textwidth]{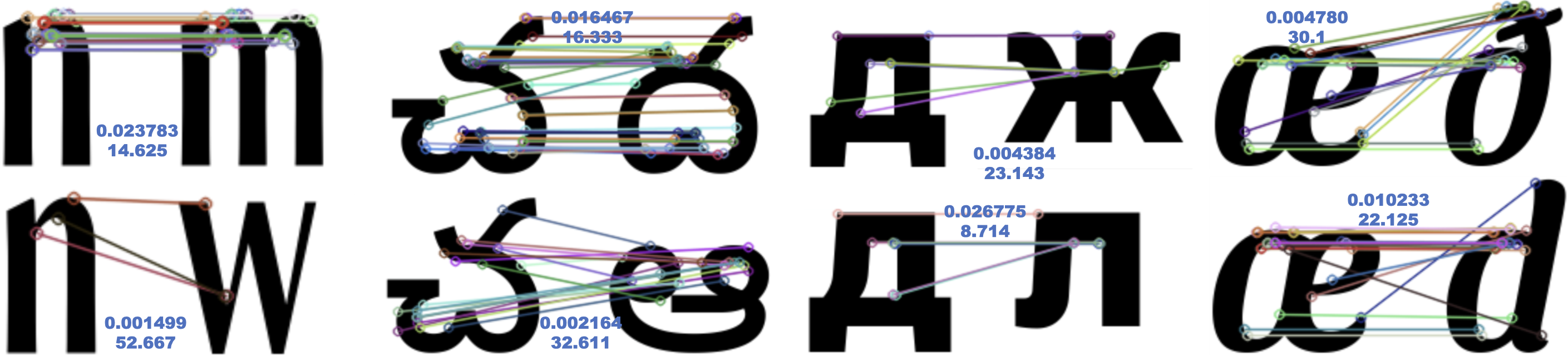}
  \caption{Visualizations of feature matching using ORB for fonts from different character sets, where matched feature points connected by colored lines. For each pair of characters, two numbers are displayed: the upper number represents the Jaccard Index $J$ of overlapping feature points, and the lower number indicates the average distance $D$.}
  \label{fig:match}
\end{figure*}

\subsection{Method $\textcircled{3}$ Real-World Injection}
\label{sec:methods_3}
Several studies have considered the generation of synthetic data through the analysis of error distributions in existing datasets. The primary approach involves embedding OCR errors into the data at rates mirroring their occurrence in real-world settings \citep{d2017generating,grundkiewicz2019neural,jasonarson2023generating}. This strategy addresses the discrepancies between existing datasets and actual application domains, offering more control over the quality of the generated text. 

To obtain a realistic OCR error distribution, in this work we use additional clean corpus $C$ coming from the same domain and apply the method described in Section \ref{sec:methods_2} to obtain OCR-processed corpus $\hat{C}$. Subsequently, the Recursive Text Alignment Scheme (RETAS) \cite{yalniz2011fast} is employed to perform the automatic alignment of the OCR text and the original clean text, allowing us to extract the probabilities for character substitution, deletion, and insertion. By adjusting the probability of these errors, we can control the CER of the synthetic data. For instance, if the CER between text $C$ and its corresponding OCR version $\hat{C}$ is \(p\), then doubling all of the error probabilities would result in an expected CER of \(2p\) for the newly generated synthetic text. In practice, we set the expected CER for each chunk in $A$ to a random value $\in [0, 15]$ to generate the synthetic data in $\hat{A}$.

\subsection{Method $\textcircled{4}$ Glyph Similarity}
\label{sec:methods_4}
Given that OCR replacement errors frequently happen between characters with visually similar glyphs \cite{jatowt2019deep}, we can also use the information from glyph similarities to construct synthetic data. Specifically, we apply the following procedure. Firstly, we filter infrequently appearing characters from the input corpus $A$ to form a definitive character set. Based on the analysis of the ICDAR2017 \cite{chiron2017icdar2017} and ICDAR2019 \cite{rigaud2019icdar} datasets, 1:1 mapping errors, where a single character is incorrectly identified as another character, accounted for 87.9\% of all 1:$n$ mapping errors. Here 1:$n$ mapping error refers to cases where a single character be incorrectly recognized as $n$ characters. To enhance the computational efficiency of the implementation proposed in this paper, we focus solely on 1:1 errors. Note that simulations of 1:$n\;(n>1)$  errors could be achieved through random insertion.

For each language, we select a set of appropriate fonts $F$, which includes a variety of historical and modern fonts, and employ a set of vision feature detection algorithms $Q$ to extract and match image features. In our experiments we consider ORB \cite{rublee2011orb}, AKAZE \cite{alcantarilla2011fast} and SIFT \cite{ng2003sift}. For each character $i$ in the character set, we calculate its glyph similarity with every other character $j$ by creating their images using fonts $f$ from the font set $F$. We then calculate their similarity under detector $q$ as:
\[S(i, j,q) =\frac{1}{|F|}  \sum_{f \in F}  \frac{J(i,j,f,q)}{D(i, j, f,q)} \] 
where $D(i, j, f, q)$ is the average distance between matching point pairs for character $i$ and $j$, using font $f$ and feature detector $q$. The degree of overlap for feature matching points $J(i, j, f, q)$ is calculated using the Jaccard Index \citep{jaccard12index}. This is defined as the ratio of the number of matching points relative to the combined number of feature points in $i$ and $j$ minus the number of matching points. A number of matching examples are shown in Figure \ref{fig:match}. We see that, as glyphs become more similar, $J$ increases and $D$ decreases.

\begin{figure}[!t]
  \centering
  \includegraphics[width=0.38\textwidth]{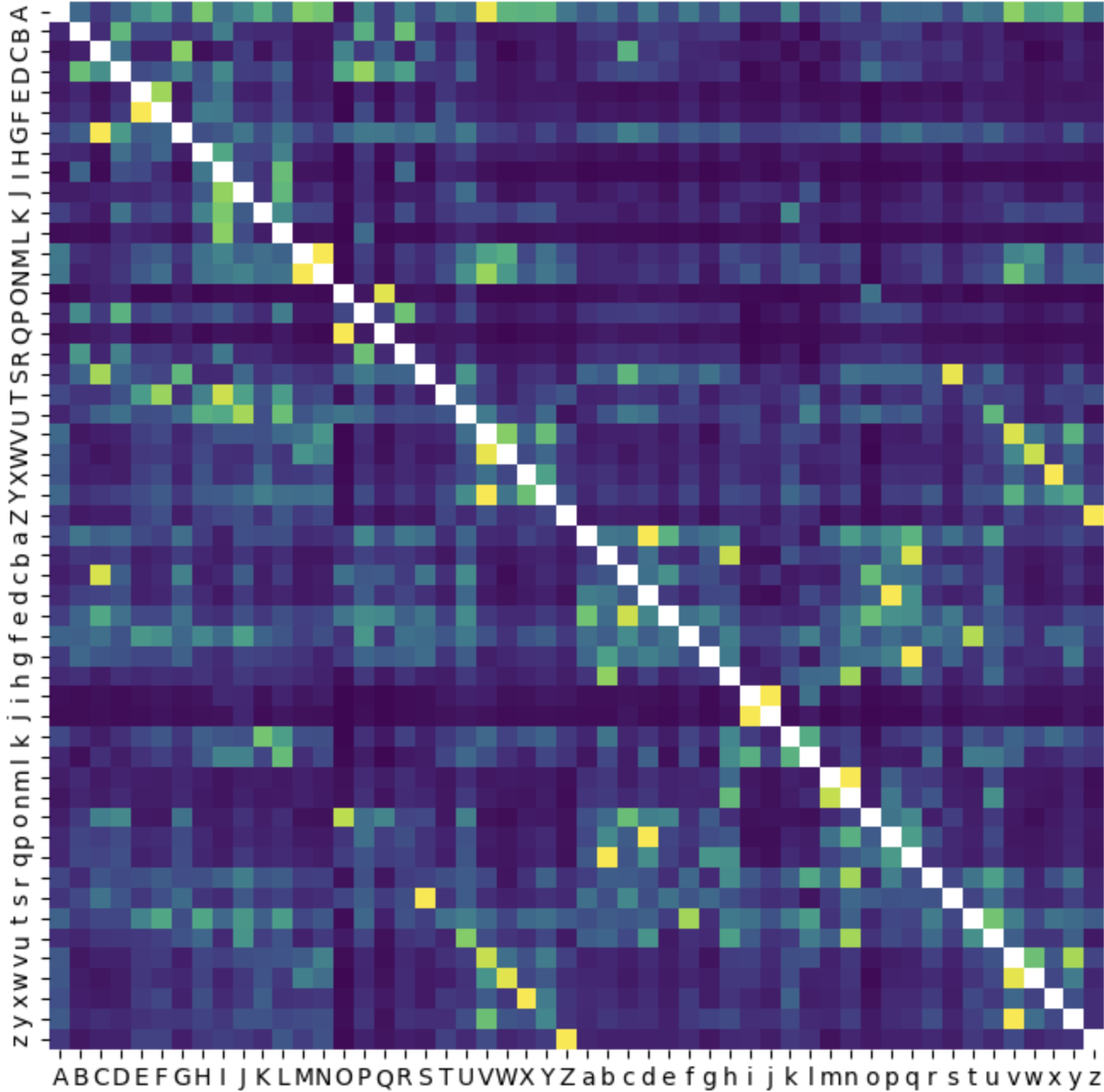}
  \caption{Visualization of a glyph similarity matrix for English-language characters (52 letters only). The saturation of each cell represents the value $S_{\text{norm}}(i, j)$ between each pair of characters $i$ and $j$.}
  \label{fig:sim}
\end{figure}

Next, we perform min-max normalization for different mapping types and characters to obtain the normalized score $\in [0,1]$
\[ S_{\text{norm}}(i, j) =\frac{1}{|Q|} \sum_{q \in Q} \frac{S(i, j, q) - \min(S_{iq})}{\max(S_{iq}) - \min(S_{iq})} \]
where $S_{iq}$ is the set of similarity scores between character $i$ and other characters using feature detector $q$.
Figure \ref{fig:sim} presents an example of $S_{\text{norm}}(i, j)$ for English-language characters. 

The technique of embedding OCR errors by exploiting glyph similarity shares some common aspects with the random injection method described previously. Firstly, we randomly select a target error rate $p \in [0, 15]$ for each chunk from $A$, where each character $i$ in $A$ has a probability of $\frac{5}{7} \times p$ to be replaced with another character. The choice of replacement is based on $S_{\text{norm}}(i, j)$, with higher-weighted pairs being more likely to be chosen. After replacement, each character has a probability of $\frac{1}{7} \times p$ to be deleted. Any two characters have a probability of $\frac{1}{7} \times p$ of having a random character inserted between them.

\section{Datasets}
In this paper, we undertake post-OCR experiments involving English, Frisian, German, Icelandic, Irish, Russian, Spanish, and Telugu texts. Clean corpora $A$ for these languages were obtained as follows: data for Icelandic, Irish, and Frisian were sourced from the CC-100 corpus \cite{conneau2019unsupervised}. Telugu\footnote{\url{https://www.kaggle.com/datasets/sudalairajkumar/telugu-nlp}} and Russian\footnote{\url{https://www.kaggle.com/datasets/d0rj3228/russian-literature/data}} datasets were obtained from Kaggle, while English, Spanish, and German were sourced from Project Gutenberg. 
Summary statistics for the corpora are given in Table \ref{tab:dataset}. The amount of data selected simulates the real-world rarity of these languages to some extent.

\begin{table}[h]
\centering
\resizebox{0.37\textwidth}{!}{
\begin{tabular}{lrl}
\toprule
\textbf{Language} & \textbf{Length} & \textbf{Source} \\
\midrule
English           & 27,883,394      & Gutenberg       \\
Frisian           & 3,426,499       & CC100           \\
German            & 3,758,352       & Gutenberg       \\
Icelandic         & 3,147,864       & CC100           \\
Irish             & 5,090,436       & CC100           \\
Russian           & 6,613,093       & Kaggle          \\
Spanish           & 7,813,245       & Gutenberg       \\
Telugu            & 5,777,551       & Kaggle          \\
\bottomrule
\end{tabular}
}
\caption{Corpus character counts and sources.}
\label{tab:dataset}
\end{table}

All training, validation, and test datasets are generated from these clean corpora. The training data is created using the four different methods described in Section \ref{sec:methods}, with the aim of comparing the synthetic data generation techniques. To implement the method presented in Section \ref{sec:methods_3}, 20\% of the text from each language corpus is used as $C$ for extracting OCR error distributions. The remaining 80\% of the corpus is divided in a 8:1:1 ratio to generate training, validation, and test sets.

The test datasets are generated using a method similar to that described in Section \ref{sec:methods_2}, employing the Tesseract OCR system \cite{smith2007overview} to convert synthetic images into OCR text.  Since the methods described in Sections \ref{sec:methods_2} and \ref{sec:methods_3} involve using an OCR system to either generate OCR text directly or to extract OCR errors, the Google Vision API OCR system \cite{fujii2017sequence} is used when creating training and validation datasets.

\section{Models}

In our experiments, we compare the performance of various models which have been previously adopted for post-OCR tasks, including models that operate at the subword-level, character-level, and byte-level tokenizers. Additionally, we evaluate current SOTA models which is based on n-gram and majority voting. This comparison encompasses both pre-trained models and those trained from scratch. Training parameters are in Appendix \ref{sec:appendix}.

\subsection{mT5}
The mT5 model \cite{xue2020mt5} is an extension of the T5 model \cite{raffel2020exploring}, which is pre-trained on a multilingual dataset. This model leverages the original T5's text-to-text framework, where every natural language processing task is reframed as a text generation problem, allowing for consistent and flexible handling of a wide range of tasks across languages. The version we use in our experiments is mT5-base.

\subsection{mBART}

The mBART model \cite{tang2020multilingual} is a multilingual extension of the BART architecture \cite{lewis2019bart}. Developed by Facebook AI, its pre-training approach involves corrupting text with an arbitrary noising function and then learning to reconstruct the original text. mBART is also pre-trained on a large corpus of text in multiple languages, making it adept at both high-resource and low-resource language translation, as well as a variety of other language processing tasks. The version we use in our experiments is mBART-large-50.

\subsection{ByT5}

The ByT5 model \cite{xue2022byt5} is designed to address the limitations of traditional subword tokenization methods by working at the byte level. This approach allows ByT5 to handle any language or writing system with Unicode representation, making it naturally multilingual and adaptable for handling a wide variety of text data. By operating on bytes, ByT5 avoids the complexities and biases associated with subword vocabularies, enabling more equitable and accurate processing across languages. ByT5 was pre-trained on the same dataset as mT5. The model demonstrates robust performance across a wide range of tasks \cite{stankevivcius2022correcting,jentoft2023grammatical}. The version we use in our experiments is ByT5-base.

\subsection{CharBERT + Glyph Embedding}

\begin{figure}[!t]
  \centering
  \includegraphics[width=0.43\textwidth]{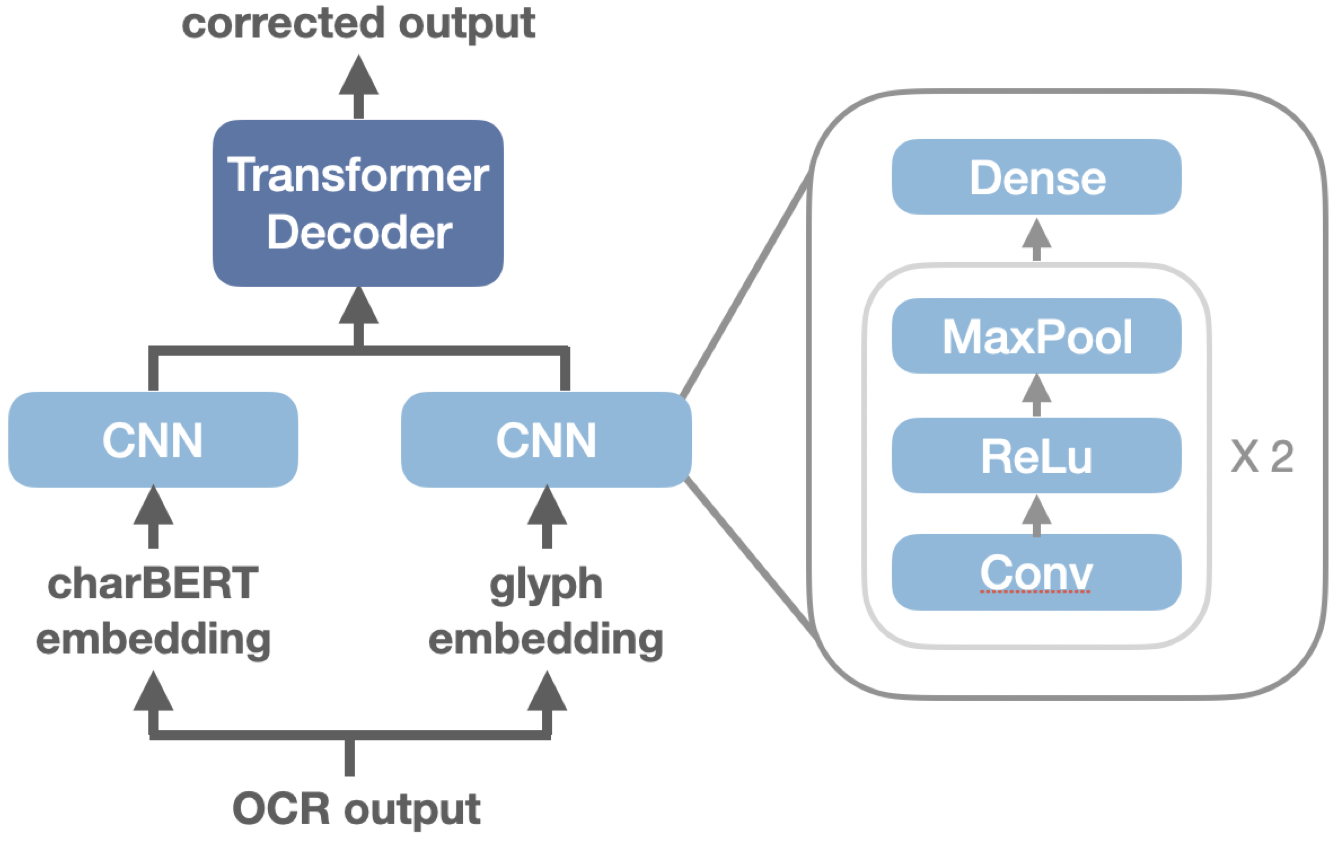}
  \vskip -0.5em
  \caption{The structure of the CharBERT post-OCR model incorporates glyph embeddings as inputs. It consists of two CNN encoders and one transformer decoder.}
  \label{fig:charbert}
\end{figure}

To verify the effectiveness of glyph embedding as proposed by \citet{chen2023enhancing}, we also conducted some experiments using their model. CharBERT \cite{ma2020charbert} is a char-level model. In the pre-training phase of CharBERT, one of the tasks involves text denoising which is closely aligns with the objectives of post-OCR correction tasks. Building on the original model, \citet{chen2023enhancing} introduced glyph embedding to CharBERT by using a ResNet50 model \cite{he2016deep}. 
This supports the extraction of visual information from characters and serves as one of the inputs to the post-OCR correction model. The structure of this model is shown in Figure \ref{fig:charbert}. In our experiments we use the same structure and fine-tuning settings corresponding to the optimal configurations reported in the original work.
Since the CharBERT model was pre-trained on the English Wikipedia dataset \cite{wikidump}, we conduct experiments only in English with this model. 

\subsection{Non-pre-trained Models}
The method proposed by \citet{ramirez2022post} achieved SOTA results in 5 out of 9 languages on the ICDAR2019 dataset. In this study, we also compare this model with other models using the configuration: window type as n-grams, window size of 60, decoding method as beam, weighting as uniform. We refer to as the ENSEMBLE model.

Additionally, we have defined a more conventional Seq2Seq model, referred to as SCRATCH. The SCRATCH model is designed with an embedding size of 512, feed-forward network embeddings of 2048, 4 attention heads, 5 encoder layers, 5 decoder layers, a SentencePiece tokenizer \cite{kudo2018sentencepiece}, and a vocabulary size of 3000.

\section{Experiments}
\label{sec:exp}

The primary objective of our empirical analysis in this paper is to explore the key factors affecting the performance of models in post-OCR tasks, aiming to identify the most appropriate settings for such tasks. We conduct three different experiments:
\begin{itemize}[noitemsep]
    \item \textit{Experiment 1:} We investigate how data volume impacts on model performance, aiming to identify a generally applicable data augmentation setting that balances performance and training time.
    \item \textit{Experiment 2:} We seek identify the best method for constructing synthetic data and the best-performing model. Using the findings from Experiment 1, we apply augmentation and generate different datasets using the four methods from Section \ref{sec:methods}, in combination with multiple post-OCR models. 
    \item \textit{Experiment 3:} We extend the best settings identified in the first two experiments to a wider set of languages for post-OCR tasks, verifying the applicability and effectiveness of these settings in a broader linguistic context.
\end{itemize}

\subsection{Experiment 1}

\begin{table*}[!t]
\centering
\resizebox{0.95\textwidth}{!}{
\begin{tabular}{l|l|cccc|cccc|cccc|cccc}
\toprule
& & \multicolumn{4}{c}{\textbf{English}} & \multicolumn{4}{c}{\textbf{Icelandic}} & \multicolumn{4}{c}{\textbf{Russian}} & \multicolumn{4}{c}{\textbf{Telugu}} \\
\cmidrule(r){3-6} \cmidrule(lr){7-10} \cmidrule(l){11-14} \cmidrule(l){15-18}
& CER& \multicolumn{4}{c}{4.96} & \multicolumn{4}{c}{10.09} & \multicolumn{4}{c}{4.13} & \multicolumn{4}{c}{34.12}\\
\midrule
Model & Data & \(\frac{1}{6}\) & \(\frac{1}{3}\) & \(\frac{2}{3}\) & 1 & \(\frac{1}{6}\) & \(\frac{1}{3}\) & \(\frac{2}{3}\) & 1 & \(\frac{1}{6}\) & \(\frac{1}{3}\) & \(\frac{2}{3}\) & 1 & \(\frac{1}{6}\) & \(\frac{1}{3}\) & \(\frac{2}{3}\) & 1\\
\midrule
\multirow{4}{*}{mT5} & 1$\times$ & 4.56 & 4.29 & 3.77 & 3.42 & 9.68 & 9.33 & 9.02 & 8.87 & 3.78 & 3.33 & 2.99 & 2.74   & 31.44 & 29.80 & 28.35 & 27.70 \\
& 2$\times$  & 4.32 & 4.10 & 3.52 & 3.24 & 9.44 & 9.12  & 8.83    & 8.49 & 3.63 & 3.15 & 2.63 & 2.51  & 29.83 & 28.69 & 27.83 & 26.89 \\
& 4$\times$ & 4.24 & 3.71 & 3.13 & 3.00 & 9.35 & 9.00 & 8.60 & 8.31 & 3.54 & 3.05 & 2.40 & \textbf{2.33}  & 29.31 & 28.45 & 27.12 & \textbf{26.31} \\
& 8$\times$  & 4.24 & 3.68 & 3.05 & \textbf{2.93} & 9.32 & 8.90  & 8.54 & \textbf{8.29}                      & 3.55 & 3.02 & 2.39 & 2.36  & 29.22 & 28.44 & 27.23 & 26.32 \\
\midrule

\multirow{4}{*}{ENSEMBLE}  & 1$\times$ & 4.84 & 4.69 & 4.51 & 4.46 & 10.18 & 10.09 & 10.00 & 9.88 & 4.09 & 4.00 & 3.90 & 3.83   & 40.23 & 38.47 & 38.02 & 37.71 \\
& 2$\times$  & 4.80 & 4.63 & 4.44 & 4.30  & 10.09 & 10.03  & 9.88 & 9.74 & 4.06 & 3.85 & 3.80 & 3.71   & 38.99 & 38.65 & 38.21 & 36.60 \\
& 4$\times$  & 4.75 & 4.53 & 4.23 & \textbf{4.19}  & 10.14  & 9.94 & 9.83 & \textbf{9.64} & 4.02 & 3.72 & 3.69 & \textbf{3.52}   & 38.78 & 38.17 & 37.43 & 35.66 \\
& 8$\times$  & 4.79 & 4.48 & 4.24 & 4.22 & 10.11 & 9.97  & 9.80 & 9.69 & 4.00 & 3.74 & 3.63 & 3.60   & 38.65 & 37.75 & 36.54 & \textbf{35.59} \\

\bottomrule
\end{tabular}
}
\caption{Results showing the effects of data volume and augmentation on mT5 and ENSEMBLE model CER on English, Icelandic, and Russian texts.}
\label{table:ex1_result}
\end{table*}

The process of generating OCR synthetic data is highly stochastic, so  the same clean text can yield different synthetic text versions when OCR errors are inserted. This variability can potentially enrich the model's learning process. In this experiment, we  investigate the impact of data augmentation techniques (by replicating the clean text data multiple times to increase the volume of data) and the quantity of original clean text data used on model performance. We examine whether increasing the diversity of data via this method can enhance the model's ability to correct OCR errors.

The experiment is conducted on English, Icelandic, Russian and Telugu texts. These languages were chosen as they include both rich-resource and low-resource languages and belong to different language families, and have different grammatical structures and character sets. This diversity makes them representative for assessing different models and methods for generating synthetic data.

To explore the impact of the original data volume, we experiment with datasets of different sizes: the full dataset size, \(\frac{2}{3}\) size, \(\frac{1}{3}\) size, and \(\frac{1}{6}\) size, as in post-OCR correction tasks, the training data often need to be from the same domain as the text need to be fixed, so the settings of 1, \(\frac{2}{3}\), \(\frac{1}{3}\), and \(\frac{1}{6}\) were designed to simulate the scenarios within different domains under one language environment with varying amounts of clean texts. 
And to examine the effect of data augmentation techniques, these will be further expanded to 1$\times$, 2$\times$, 4$\times$ and 8$\times$ their original sizes. This approach will yield multiple versions of dataset \(A\), and since Method \(\textcircled{3}\) is the most commonly used, we will employ it to generate the training data for this experiment. The experiments are conducted using both the pre-trained mT5-base model and ENSEMBLE model. 

The results in Table \ref{table:ex1_result} show that mT5 outperforms the ENSEMBLE model across the four languages. Data augmentation enhances model performance with diminishing returns -- augmenting from 1$\times$ to 4$\times$ effectively lowers CER, but gains from 4$\times$ to 8$\times$ are marginal. In our experiment, training parameters for mT5 included a learning rate of 5e-4, warm-up steps of 250, batch size of 4, dropout rate of 0.2, and 6 epochs on fp32. Training times for 4$\times$ augmented datasets on an RTX 4090 were approximately 35, 5, 10 and 8 hours for English, Icelandic, Russian and Telugu, respectively. Overall, the results suggest that 4$\times$ augmentation provides sufficient improvement.

\subsection{Experiment 2}

Next, we explore how different methods of creating synthetic data impact on the performance of different language models. We conduct tests using several popular models: mT5-base \cite{xue2020mt5}, ByT5-base \cite{xue2022byt5}, mBART-large \cite{chipman2022mbart}, SCRATCH, and ENSEMBLE. We also evaluate the approach of \citet{chen2023enhancing}, which combines CharBERT \cite{ma2020charbert} with glyph embedding, and compare these results to CharBERT without glyph embedding.

\begin{table*}[!t]
\centering
\resizebox{0.95\textwidth}{!}{
\begin{tabular}{l|cccc|cccc|cccc|cccc}
\toprule
& \multicolumn{4}{c}{\textbf{English}} & \multicolumn{4}{c}{\textbf{Icelandic}} & \multicolumn{4}{c}{\textbf{Russian}} & \multicolumn{4}{c}{\textbf{Telugu}} \\
\cmidrule(r){2-5} \cmidrule(lr){6-9} \cmidrule(l){10-13} \cmidrule(l){14-17}
CER& \multicolumn{4}{c}{4.96} & \multicolumn{4}{c}{10.09} & \multicolumn{4}{c}{4.13} & \multicolumn{4}{c}{34.12}\\
\midrule
Method & \textcircled{1} & \textcircled{2} & \textcircled{3} & \textcircled{4} & \textcircled{1} & \textcircled{2} & \textcircled{3} & \textcircled{4} & \textcircled{1} & \textcircled{2} & \textcircled{3} & \textcircled{4} & \textcircled{1} & \textcircled{2} & \textcircled{3} & \textcircled{4}\\
\midrule
mT5 & \textbf{3.98} & 3.42 & \uline{3.00} & 3.03 & 9.48 & 9.53 & \uline{\textbf{8.31}} & 8.38 & 3.36 & \textbf{2.98} & \textbf{2.33} & \uline{2.27}  & 30.35 & 28.24 & 26.31 & \uline{25.85} \\
ByT5 & 4.02 & \textbf{3.36} & \textbf{\uline{2.96}} & \textbf{3.00} & \textbf{9.42} & \textbf{9.45} & 8.39 & \textbf{\uline{8.28}} & \textbf{3.34} & 3.13 & 2.34 & \textbf{\uline{2.14}} & \textbf{30.21} & \textbf{28.14} & \textbf{25.98} & \textbf{\uline{25.28}} \\
mBART & 4.19 & 4.23 & 3.63 & \uline{3.54} & 9.93 & 10.26 & 9.41 & \uline{9.30} & 3.41 & 3.33 & 2.82 & \uline{2.78} & 31.06 & 28.49 & \uline{25.66} & 26.00\\
CharBERT & 4.12 & 3.60 & \uline{3.39} & 3.57 & - & - & - & - & - & - & - & - & - & - & - & - \\
\quad+Glyph & 4.11 & 3.61 & \uline{3.42} & 3.54 & - & - & - & - & - & - & - & - & - & - & - & -\\

ENSEMBLE & 4.56 & 4.44 & 4.19 & \uline{4.00} & 10.32 & 10.65  & 9.64 & \uline{9.54} & 3.77 & 3.83 & 3.52 & \uline{3.34}  & 37.12 & 34.23 & \uline{35.66} & 37.50\\
SCRATCH  & 4.79 & 4.74 & \uline{4.52} & 4.62 & 11.92 & \uline{9.82}  & 10.43 & 10.00 & 3.91 & 4.06 & \uline{3.62} & 3.64  & 33.56 & 35.28 & \uline{34.52} & 34.92\\
\bottomrule
\end{tabular}
}
\caption{Comparison of CER results across various models and synthetic data generation methods in English, Icelandic, and Russian. The best-performing model for each dataset is highlighted in bold, and the best method for generating synthetic data for each model is underlined.}
\label{table:ex2_result}
\end{table*}

\begin{table*}[!t]
\centering
\resizebox{0.99\textwidth}{!}{
\centering
\begin{tabular}{l|rr|rr|rr|rr|rr|rr|rr|rr}
\toprule
\multirow{2}{*}{\textbf{}}  & \multicolumn{2}{c|}{\textbf{English}} & \multicolumn{2}{c|}{\textbf{German}}  & \multicolumn{2}{c|}{\textbf{Irish}} & \multicolumn{2}{c|}{\textbf{Icelandic}}  & \multicolumn{2}{c|}{\textbf{Frisian}}& \multicolumn{2}{c|}{\textbf{Russian}} & \multicolumn{2}{c|}{\textbf{Spanish}} & \multicolumn{2}{c}{\textbf{Telugu}} \\ 
\cmidrule(lr){2-3} \cmidrule(lr){4-5} \cmidrule(lr){6-7} \cmidrule(lr){8-9} \cmidrule(lr){10-11} \cmidrule(lr){12-13} \cmidrule(lr){14-15} \cmidrule(lr){16-17}   
 & \textbf{CER} & \textbf{WER} & \textbf{CER} & \textbf{WER} & \textbf{CER} & \textbf{WER} & \textbf{CER} & \textbf{WER} & \textbf{CER} & \textbf{WER} & \textbf{CER} & \textbf{WER} & \textbf{CER} & \textbf{WER} & \textbf{CER} & \textbf{WER} \\ 
\midrule
OCR      & 4.96 & 15.43 & 5.79 & 19.29 & 12.57 & 35.99 & 10.09 & 30.06 & 5.15 & 16.20 & 4.13 & 8.16 & 6.00 & 17.38 & 34.12 & 90.41\\ \hline
Post-OCR & 3.00 & 7.74  & 4.27 & 11.07 & 11.01 & 29.97 & 8.28  & 24.57 & 3.55 & 11.03 & 2.14 & 5.88 & 3.76 & 8.92  & 25.28 & 66.64 \\ 
\bottomrule
\end{tabular}%
}
\caption{Performance of the ByT5 model in terms of Character Error Rate (CER) and Word Error Rate (WER) across multiple languages, before and after post-OCR correction.}
\label{tab:ex3_result}
\end{table*}

The results of Experiment 2 are shown in Table \ref{table:ex2_result}. It is observed that methods \(\textcircled{3}\) and \(\textcircled{4}\) generally outperform \(\textcircled{1}\) and \(\textcircled{2}\). Of the latter two, method \(\textcircled{1}\), which does not rely on additional data, yields slightly higher accuracy. While method \(\textcircled{2}\) allows the model to learn from real OCR outputs, it fails to correct sentences with varying degrees of CER. In our experiments, the Google Vision API OCR system was used to generate training data, and Tesseract was used for test data. Since the  Vision API OCR system generally has a higher accuracy than Tesseract, models only learn to correct sentences with low CER, which is reflected in the limitations of the method \(\textcircled{2}\). 

Method \(\textcircled{3}\) proves more effective, extracting real OCR error distributions and generating training data with various CER values, but requires additional data. 
With a sufficient volume of data (as in the English language experiments), models trained with this method performed best. Different OCR systems, due to their unique designs, can make various kinds of errors. With insufficient additional data, Method \(\textcircled{3}\) can fail to extract a comprehensive set of OCR errors, leading to distributional shift in the data, which ultimately affects the performance of models that rely on it. In Experiment 2, Method \(\textcircled{4}\) achieved impressive results without relying on additional data. By generating its own training data with varying degrees of CER encoded through glyphs alone, it successfully trained models that perform well across diverse languages with different character sets. 

Overall, we observe that pre-trained models significantly outperformed smaller Seq2Seq models trained from scratch. mT5 and ByT5 showed similar performance, with ByT5 reducing CER values by 39.5\%, 17.9\%, 48.2\% and 25.9\% on English, Icelandic, Russian and Telugu datasets, respectively. CharBERT did not benefit from glyph embedding,  consistent with the findings of \citet{amrhein2018supervised}. 


\subsection{Experiment 3}

In our final experiment, we replicate the original clean text four times for augmentation. Based on our previous experiments, we select Method \(\textcircled{4}\) to generate synthetic data and train the ByT5 model for post-OCR tasks on data from a wider set of languages, so that we can evaluate the chosen configuration in a broader linguistic context.

\begin{figure}[!h]
  \centering
  \includegraphics[width=0.48\textwidth]{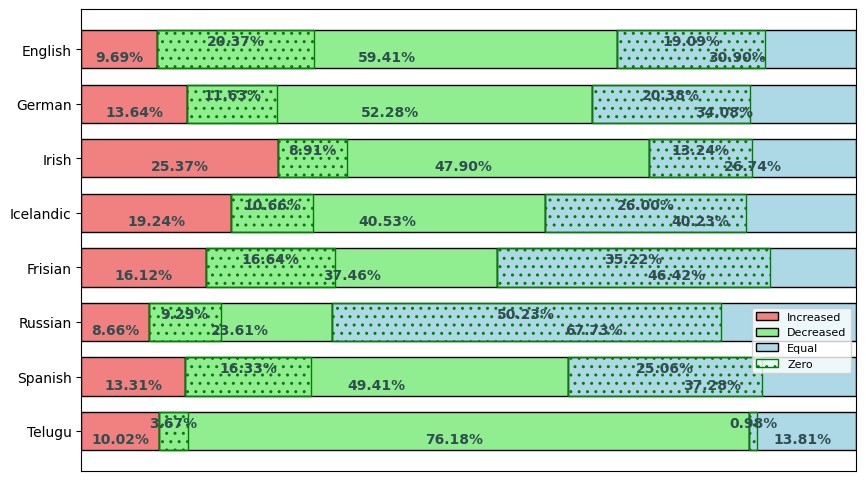}
  \caption{Comparative analysis of CER changes after post-OCR across multiple languages. Each bar represents the distribution of CER changes categorized as Increased (red), Decreased (green), Equal (blue), and Zero (dotted green).}
  \label{fig:result}
\end{figure}

From the results in Table \ref{tab:ex3_result}, we see that the model performs well on the English, Russian, and Spanish tasks, with CER reductions of 39.52\%, 48.18\%, and 37.33\%, respectively. This approach also achieves a CER reduction of 31.07\% on the low-resource language Frisian. The performance is poorest in the cases of Irish and Icelandic, with only 12.41\% and 17.94\% reductions in CER, respectively. We attribute this to two main reasons. Firstly, the data for Irish and Icelandic (sourced from the CC100 dataset of web-crawled data) contain numerous proper nouns. These are challenging for conventional NMT models to correct due to their inability to handle out-of-vocabulary words. Secondly, although the ByT5 model was trained on a multilingual dataset, the volume of data for low-resource languages is not substantial. This disparity in data availability can affect the models' performance on post-OCR tasks.  Table \ref{tab:ex3_result} also indicates that the Telugu test dataset has very high CER and WER, likely due to the complexity of Telugu glyphs \citep{negi2001ocr}. When Telugu text becomes blurred, it can be difficult for OCR systems to recognize. Thus, post-OCR processing for Telugu results in a 25.91\% reduction in CER. 

Figure \ref{fig:result} provides a more detailed sentence-based view of CER across languages. Here we enumerate the following categories for CER changes after post-OCR processing: "Increased" indicates sentences where the CER has increased, "Decreased" indicates the CER has decreased, "Equal" means that CER remained unchanged, and "Zero" means the CER is zero after post-OCR processing (i.e., perfect matching with the ground truth). We observe that only a small number of sentences experienced an increase in CER after post-OCR processing. Specifically, 25.37\% of sentences in Irish showed an increase in CER post-OCR, and only 8.66\% of sentences in Russian. Notably, 59.52\% of sentences in Russian had a CER of zero after post-OCR processing. For Telugu, which had the highest CER, we saw a reduction in CER for 76.18\% of sentences after post-OCR processing.

\section{Conclusion}

Our experiments demonstrated that augmenting data by replicating clean text and constructing synthetic data based on glyph similarity significantly improves model performance for post-OCR correction. Furthermore, we proposed a novel approach to constructing post-OCR synthetic data based on glyph similarity, which outperforms traditional noise injection methods, especially in scenarios with limited data volume, as it does not require additional external data. In our experiments, pre-trained models demonstrated superior performance compared to those without pre-training, with the byte-level ByT5 model achieving the best performance. For rich-resource languages, this approach can reduce the CER by approximately 45\%, while for low-resource languages the reduction varies between 12.41\% and 31.07\%. 

\section*{Limitations}
For the preferred Method \textcircled{4}, the time complexity of calculating the character similarity values is $O(n^2)$. As a result, the calculation of character similarities in languages with a large array of characters, such as Chinese and Japanese, can be quite time consuming.


\section*{Acknowledgements}
This publication is part of a project that has received funding from (i) the European Research Council (ERC) under the Horizon 2020 research and innovation programme (Grant agreement No. 884951); (ii) Science Foundation Ireland (SFI) to the Insight Centre for Data Analytics under grant No 12/RC/2289 P2.

\bibliography{anthology,custom}

\begin{thebibliography}{67}
\expandafter\ifx\csname natexlab\endcsname\relax\def\natexlab#1{#1}\fi

\bibitem[{Ahmadi et~al.(2023)Ahmadi, Agarwal, and Anastasopoulos}]{ahmadi2023pali}
Sina Ahmadi, Milind Agarwal, and Antonios Anastasopoulos. 2023.
\newblock Pali: A language identification benchmark for {Perso-Arabic} scripts.
\newblock \emph{arXiv preprint arXiv:2304.01322}.

\bibitem[{Alcantarilla and Solutions(2011)}]{alcantarilla2011fast}
Pablo~F Alcantarilla and T~Solutions. 2011.
\newblock Fast explicit diffusion for accelerated features in nonlinear scale spaces.
\newblock \emph{IEEE Trans. Patt. Anal. Mach. Intell}, 34(7):1281--1298.

\bibitem[{Amrhein and Clematide(2018)}]{amrhein2018supervised}
Chantal Amrhein and Simon Clematide. 2018.
\newblock Supervised {OCR} error detection and correction using statistical and neural machine translation methods.
\newblock \emph{Journal for Language Technology and Computational Linguistics}, 33(1):49--76.

\bibitem[{Bassil and Alwani(2012)}]{bassil2012ocr}
Youssef Bassil and Mohammad Alwani. 2012.
\newblock {OCR} post-processing error correction algorithm using google online spelling suggestion.
\newblock \emph{arXiv preprint arXiv:1204.0191}.

\bibitem[{Bazzo et~al.(2020)Bazzo, Lorentz, Suarez~Vargas, and Moreira}]{bazzo2020assessing}
Guilherme~Torresan Bazzo, Gustavo~Acauan Lorentz, Danny Suarez~Vargas, and Viviane~P Moreira. 2020.
\newblock Assessing the impact of {OCR} errors in information retrieval.
\newblock In \emph{Advances in Information Retrieval: 42nd European Conference on IR Research (ECIR 2020)}, pages 102--109. Springer.

\bibitem[{Boros et~al.(2022)Boros, Nguyen, Lejeune, and Doucet}]{boros2022assessing}
Emanuela Boros, Nhu~Khoa Nguyen, Ga{\"e}l Lejeune, and Antoine Doucet. 2022.
\newblock Assessing the impact of ocr noise on multilingual event detection over digitised documents.
\newblock \emph{International Journal on Digital Libraries}, 23(3):241--266.

\bibitem[{Bubeck et~al.(2023)Bubeck, Chandrasekaran, Eldan, Gehrke, Horvitz, Kamar, Lee, Lee, Li, Lundberg et~al.}]{bubeck2023sparks}
S{\'e}bastien Bubeck, Varun Chandrasekaran, Ronen Eldan, Johannes Gehrke, Eric Horvitz, Ece Kamar, Peter Lee, Yin~Tat Lee, Yuanzhi Li, Scott Lundberg, et~al. 2023.
\newblock Sparks of artificial general intelligence: Early experiments with {GPT-4}.
\newblock \emph{arXiv preprint}, (2303.12712).

\bibitem[{Chen and Zhou(2023)}]{chen2023enhancing}
Yung-Hsin Chen and Yuli Zhou. 2023.
\newblock Enhancing ocr performance through {Post-OCR} models: Adopting glyph embedding for improved correction.
\newblock \emph{arXiv preprint arXiv:2308.15262}.

\bibitem[{Chipman et~al.(2022)Chipman, George, McCulloch, and Shively}]{chipman2022mbart}
Hugh~A Chipman, Edward~I George, Robert~E McCulloch, and Thomas~S Shively. 2022.
\newblock mbart: multidimensional monotone bart.
\newblock \emph{Bayesian Analysis}, 17(2):515--544.

\bibitem[{Chiron et~al.(2017)Chiron, Doucet, Coustaty, and Moreux}]{chiron2017icdar2017}
Guillaume Chiron, Antoine Doucet, Micka{\"e}l Coustaty, and Jean-Philippe Moreux. 2017.
\newblock {ICDAR-2017} competition on post-{OCR} text correction.
\newblock In \emph{Proceedings of the 14th IAPR International Conference on Document Analysis and Recognition (ICDAR)}, volume~1, pages 1423--1428. IEEE.

\bibitem[{Choi and Park(2023)}]{choi2023dmops}
Eujeong Choi and Chanjun Park. 2023.
\newblock Dmops: Data management operation and recipes.
\newblock \emph{arXiv preprint arXiv:2301.01228}.

\bibitem[{Clematide et~al.(2016)Clematide, Furrer, and Volk}]{clematide2016crowdsourcing}
Simon Clematide, Lenz Furrer, and Martin Volk. 2016.
\newblock Crowdsourcing an {OCR} gold standard for a german and french heritage corpus.
\newblock In \emph{Proceedings of the Tenth International Conference on Language Resources and Evaluation (LREC'16)}, pages 975--982.

\bibitem[{Conneau et~al.(2019)Conneau, Khandelwal, Goyal, Chaudhary, Wenzek, Guzm{\'a}n, Grave, Ott, Zettlemoyer, and Stoyanov}]{conneau2019unsupervised}
Alexis Conneau, Kartikay Khandelwal, Naman Goyal, Vishrav Chaudhary, Guillaume Wenzek, Francisco Guzm{\'a}n, Edouard Grave, Myle Ott, Luke Zettlemoyer, and Veselin Stoyanov. 2019.
\newblock Unsupervised cross-lingual representation learning at scale.
\newblock \emph{arXiv preprint arXiv:1911.02116}.

\bibitem[{Davydkin et~al.(2023)Davydkin, Markelov, Iuldashev, Dudkin, and Krivorotov}]{davydkin2023data}
Evgenii Davydkin, Aleksandr Markelov, Egor Iuldashev, Anton Dudkin, and Ivan Krivorotov. 2023.
\newblock \href {http://arxiv.org/abs/2311.15896} {Data generation for post-ocr correction of cyrillic handwriting}.

\bibitem[{Devlin et~al.(2018)Devlin, Chang, Lee, and Toutanova}]{devlin2018bert}
Jacob Devlin, Ming-Wei Chang, Kenton Lee, and Kristina Toutanova. 2018.
\newblock Bert: Pre-training of deep bidirectional transformers for language understanding.
\newblock \emph{arXiv preprint}, (1810.04805).

\bibitem[{D’hondt et~al.(2017)D’hondt, Grouin, and Grau}]{d2017generating}
Eva D’hondt, Cyril Grouin, and Brigitte Grau. 2017.
\newblock Generating a training corpus for {OCR} post-correction using encoder-decoder model.
\newblock In \emph{Proceedings of the 8th International Joint Conference on Natural Language Processing (Volume 1: Long Papers)}, pages 1006--1014.

\bibitem[{Estrella and Paliza(2014)}]{estrella2014ocr}
Paula Estrella and Pablo Paliza. 2014.
\newblock Ocr correction of documents generated during argentina's national reorganization process.
\newblock In \emph{Proceedings of the First International Conference on Digital Access to Textual Cultural Heritage}, pages 119--123.

\bibitem[{Foundation(2024)}]{wikidump}
Wikimedia Foundation. 2024.
\newblock \href {https://dumps.wikimedia.org} {Wikimedia downloads}.

\bibitem[{Fujii et~al.(2017)Fujii, Driesen, Baccash, Hurst, and Popat}]{fujii2017sequence}
Yasuhisa Fujii, Karel Driesen, Jonathan Baccash, Ash Hurst, and Ashok~C Popat. 2017.
\newblock Sequence-to-label script identification for multilingual {OCR}.
\newblock In \emph{2017 14th IAPR international conference on document analysis and recognition (ICDAR)}, volume~1, pages 161--168. IEEE.

\bibitem[{Furrer and Volk(2011)}]{furrer2011reducing}
Lenz Furrer and Martin Volk. 2011.
\newblock Reducing ocr errors in gothic-script documents.
\newblock In \emph{Proceedings of the Workshop on Language Technologies for Digital Humanities and Cultural Heritage}, pages 97--103.

\bibitem[{Ginsberg and Yu(2018)}]{ginsberg2018rapid}
Avi Ginsberg and Cui Yu. 2018.
\newblock Rapid homoglyph prediction and detection.
\newblock In \emph{Proc. 1st International Conference on Data Intelligence and Security (ICDIS)}, pages 17--23. IEEE.

\bibitem[{Grundkiewicz et~al.(2019)Grundkiewicz, Junczys-Dowmunt, and Heafield}]{grundkiewicz2019neural}
Roman Grundkiewicz, Marcin Junczys-Dowmunt, and Kenneth Heafield. 2019.
\newblock Neural grammatical error correction systems with unsupervised pre-training on synthetic data.
\newblock In \emph{Proceedings of the Fourteenth Workshop on Innovative Use of NLP for Building Educational Applications}, pages 252--263.

\bibitem[{He et~al.(2016)He, Zhang, Ren, and Sun}]{he2016deep}
Kaiming He, Xiangyu Zhang, Shaoqing Ren, and Jian Sun. 2016.
\newblock Deep residual learning for image recognition.
\newblock In \emph{Proceedings of the IEEE conference on computer vision and pattern recognition}, pages 770--778.

\bibitem[{Ignat et~al.(2022)Ignat, Maillard, Chaudhary, and Guzm{\'a}n}]{ignat2022ocr}
Oana Ignat, Jean Maillard, Vishrav Chaudhary, and Francisco Guzm{\'a}n. 2022.
\newblock {OCR} improves machine translation for low-resource languages.
\newblock \emph{arXiv preprint arXiv:2202.13274}.

\bibitem[{Ing{\'o}lfsd{\'o}ttir et~al.(2023)Ing{\'o}lfsd{\'o}ttir, Ragnarsson, J{\'o}nsson, S{\'\i}monarson, {\TH}orsteinsson, and Sn{\ae}bjarnarson}]{ingolfsdottir2023byte}
Svanhv{\'\i}t~Lilja Ing{\'o}lfsd{\'o}ttir, P{\'e}tur~Orri Ragnarsson, Haukur~P{\'a}ll J{\'o}nsson, Haukur~Barri S{\'\i}monarson, Vilhj{\'a}lmur {\TH}orsteinsson, and V{\'e}steinn Sn{\ae}bjarnarson. 2023.
\newblock Byte-level grammatical error correction using synthetic and curated corpora.
\newblock \emph{arXiv preprint arXiv:2305.17906}.

\bibitem[{Izumi et~al.(2003)Izumi, Uchimoto, Saiga, Supnithi, and Isahara}]{izumi2003automatic}
Emi Izumi, Kiyotaka Uchimoto, Toyomi Saiga, Thepchai Supnithi, and Hitoshi Isahara. 2003.
\newblock Automatic error detection in the japanese learners’ english spoken data.
\newblock In \emph{The Companion Volume to the Proceedings of 41st Annual Meeting of the Association for Computational Linguistics}, pages 145--148.

\bibitem[{Jaccard(1912)}]{jaccard12index}
Paul Jaccard. 1912.
\newblock The distribution of flora in the alpine zone.
\newblock \emph{New Phytologist}, 11(2):37--50.

\bibitem[{Jaderberg et~al.(2014)Jaderberg, Simonyan, Vedaldi, and Zisserman}]{jaderberg2014synthetic}
Max Jaderberg, Karen Simonyan, Andrea Vedaldi, and Andrew Zisserman. 2014.
\newblock Synthetic data and artificial neural networks for natural scene text recognition.
\newblock \emph{arXiv preprint arXiv:1406.2227}.

\bibitem[{Jarlbrink and Snickars(2017)}]{jarlbrink2017cultural}
Johan Jarlbrink and Pelle Snickars. 2017.
\newblock Cultural heritage as digital noise: nineteenth century newspapers in the digital archive.
\newblock \emph{Journal of Documentation}, 73(6):1228--1243.

\bibitem[{Jasonarson et~al.(2023)Jasonarson, Steingr{\'\i}msson, Sigur{\dh}sson, Magn{\'u}sson, and Ingimundarson}]{jasonarson2023generating}
Atli Jasonarson, Stein{\th}{\'o}r Steingr{\'\i}msson, Einar~Freyr Sigur{\dh}sson, {\'A}rni~Dav{\'\i}{\dh} Magn{\'u}sson, and Finnur~{\'A}g{\'u}st Ingimundarson. 2023.
\newblock {Generating Errors: {OCR} Post-Processing for {Icelandic}}.
\newblock In \emph{24th Nordic Conference on Computational Linguistics}.

\bibitem[{Jatowt et~al.(2019)Jatowt, Coustaty, Nguyen, Doucet et~al.}]{jatowt2019deep}
Adam Jatowt, Mickael Coustaty, Nhu-Van Nguyen, Antoine Doucet, et~al. 2019.
\newblock Deep statistical analysis of {OCR} errors for effective post-{OCR} processing.
\newblock In \emph{2019 ACM/IEEE Joint Conference on Digital Libraries (JCDL)}, pages 29--38. IEEE.

\bibitem[{Jentoft(2023)}]{jentoft2023grammatical}
Matias Jentoft. 2023.
\newblock Grammatical error correction with byte-level language models.
\newblock Master's thesis, University of Oslo.

\bibitem[{Kettunen(2016)}]{kettunen2016keep}
Kimmo Kettunen. 2016.
\newblock Keep, change or delete? setting up a low resource ocr post-correction framework for a digitized old finnish newspaper collection.
\newblock In \emph{Digital Libraries on the Move: 11th Italian Research Conference on Digital Libraries, IRCDL 2015, Bolzano, Italy, January 29-30, 2015, Revised Selected Papers 11}, pages 95--103. Springer.

\bibitem[{Koehn et~al.(2020)Koehn, Chaudhary, El-Kishky, Goyal, Chen, and Guzm{\'a}n}]{koehn2020findings}
Philipp Koehn, Vishrav Chaudhary, Ahmed El-Kishky, Naman Goyal, Peng-Jen Chen, and Francisco Guzm{\'a}n. 2020.
\newblock Findings of the wmt 2020 shared task on parallel corpus filtering and alignment.
\newblock In \emph{Proceedings of the Fifth Conference on Machine Translation}, pages 726--742.

\bibitem[{Koudoro-Parfait et~al.(2021)Koudoro-Parfait, Lejeune, and Roe}]{koudoro2021spatial}
Caroline Koudoro-Parfait, Ga{\"e}l Lejeune, and Glenn Roe. 2021.
\newblock Spatial named entity recognition in literary texts: What is the influence of ocr noise?
\newblock In \emph{Proceedings of the 5th ACM SIGSPATIAL International Workshop on Geospatial Humanities}, pages 13--21.

\bibitem[{Krishna et~al.(2018)Krishna, Majumder, Bhat, and Goyal}]{krishna2018upcycle}
Amrith Krishna, Bodhisattwa~Prasad Majumder, Rajesh~Shreedhar Bhat, and Pawan Goyal. 2018.
\newblock Upcycle your ocr: Reusing ocrs for post-ocr text correction in romanised sanskrit.
\newblock \emph{arXiv preprint arXiv:1809.02147}.

\bibitem[{Kudo and Richardson(2018)}]{kudo2018sentencepiece}
Taku Kudo and John Richardson. 2018.
\newblock Sentencepiece: A simple and language independent subword tokenizer and detokenizer for neural text processing.
\newblock \emph{arXiv preprint}, (1808.06226).

\bibitem[{Lam and Suen(1997)}]{lam1997application}
Louisa Lam and SY~Suen. 1997.
\newblock Application of majority voting to pattern recognition: an analysis of its behavior and performance.
\newblock \emph{IEEE Transactions on Systems, Man, and Cybernetics-Part A: Systems and Humans}, 27(5):553--568.

\bibitem[{Lewis et~al.(2019)Lewis, Liu, Goyal, Ghazvininejad, Mohamed, Levy, Stoyanov, and Zettlemoyer}]{lewis2019bart}
Mike Lewis, Yinhan Liu, Naman Goyal, Marjan Ghazvininejad, Abdelrahman Mohamed, Omer Levy, Ves Stoyanov, and Luke Zettlemoyer. 2019.
\newblock Bart: Denoising sequence-to-sequence pre-training for natural language generation, translation, and comprehension.
\newblock \emph{arXiv preprint arXiv:1910.13461}.

\bibitem[{Li et~al.(2022)Li, Hou, and Che}]{li2022data}
Bohan Li, Yutai Hou, and Wanxiang Che. 2022.
\newblock Data augmentation approaches in natural language processing: A survey.
\newblock \emph{Ai Open}, 3:71--90.

\bibitem[{Linhares~Pontes et~al.(2019)Linhares~Pontes, Hamdi, Sidere, and Doucet}]{linhares2019impact}
Elvys Linhares~Pontes, Ahmed Hamdi, Nicolas Sidere, and Antoine Doucet. 2019.
\newblock Impact of ocr quality on named entity linking.
\newblock In \emph{Digital Libraries at the Crossroads of Digital Information for the Future: 21st International Conference on Asia-Pacific Digital Libraries, ICADL 2019, Kuala Lumpur, Malaysia, November 4--7, 2019, Proceedings 21}, pages 102--115. Springer.

\bibitem[{Liu et~al.(2021)Liu, Ding, Bing, Joty, Si, and Miao}]{liu2021mulda}
Linlin Liu, Bosheng Ding, Lidong Bing, Shafiq Joty, Luo Si, and Chunyan Miao. 2021.
\newblock {MulDA}: A multilingual data augmentation framework for low-resource cross-lingual {NER}.
\newblock In \emph{Proceedings of the 59th Annual Meeting of the Association for Computational Linguistics and the 11th International Joint Conference on Natural Language Processing (Volume 1: Long Papers)}, pages 5834--5846.

\bibitem[{Ma et~al.(2020)Ma, Cui, Si, Liu, Wang, and Hu}]{ma2020charbert}
Wentao Ma, Yiming Cui, Chenglei Si, Ting Liu, Shijin Wang, and Guoping Hu. 2020.
\newblock Charbert: character-aware pre-trained language model.
\newblock \emph{arXiv preprint arXiv:2011.01513}.

\bibitem[{Maheshwari et~al.(2022)Maheshwari, Singh, Krishna, and Ramakrishnan}]{maheshwari2022benchmark}
Ayush Maheshwari, Nikhil Singh, Amrith Krishna, and Ganesh Ramakrishnan. 2022.
\newblock A benchmark and dataset for post-{OCR} text correction in {Sanskrit}.
\newblock \emph{arXiv preprint arXiv:2211.07980}.

\bibitem[{Mazumder et~al.(2022)Mazumder, Banbury, Yao, Karla{\v{s}}, Rojas, Diamos, Diamos, He, Parrish, Kirk et~al.}]{mazumder2022dataperf}
Mark Mazumder, Colby Banbury, Xiaozhe Yao, Bojan Karla{\v{s}}, William~Gaviria Rojas, Sudnya Diamos, Greg Diamos, Lynn He, Alicia Parrish, Hannah~Rose Kirk, et~al. 2022.
\newblock Dataperf: Benchmarks for data-centric ai development.
\newblock \emph{arXiv preprint arXiv:2207.10062}.

\bibitem[{Negi et~al.(2001)Negi, Bhagvati, and Krishna}]{negi2001ocr}
Atul Negi, Chakravarthy Bhagvati, and B~Krishna. 2001.
\newblock An {OCR} system for {Telugu}.
\newblock In \emph{Proceedings of 6th International Conference on Document Analysis and Recognition}, pages 1110--1114. IEEE.

\bibitem[{Ng and Henikoff(2003)}]{ng2003sift}
Pauline~C Ng and Steven Henikoff. 2003.
\newblock Sift: Predicting amino acid changes that affect protein function.
\newblock \emph{Nucleic acids research}, 31(13):3812--3814.

\bibitem[{Nguyen et~al.(2020)Nguyen, Jatowt, Nguyen, Coustaty, and Doucet}]{nguyen2020neural}
Thi Tuyet~Hai Nguyen, Adam Jatowt, Nhu-Van Nguyen, Mickael Coustaty, and Antoine Doucet. 2020.
\newblock Neural machine translation with {BERT} for post-{OCR} error detection and correction.
\newblock In \emph{Proceedings of the ACM/IEEE joint conference on digital libraries}, pages 333--336.

\bibitem[{Puri et~al.(2020)Puri, Spring, Patwary, Shoeybi, and Catanzaro}]{puri2020training}
Raul Puri, Ryan Spring, Mostofa Patwary, Mohammad Shoeybi, and Bryan Catanzaro. 2020.
\newblock Training question answering models from synthetic data.
\newblock \emph{arXiv preprint arXiv:2002.09599}.

\bibitem[{Raffel et~al.(2020)Raffel, Shazeer, Roberts, Lee, Narang, Matena, Zhou, Li, and Liu}]{raffel2020exploring}
Colin Raffel, Noam Shazeer, Adam Roberts, Katherine Lee, Sharan Narang, Michael Matena, Yanqi Zhou, Wei Li, and Peter~J Liu. 2020.
\newblock Exploring the limits of transfer learning with a unified text-to-text transformer.
\newblock \emph{The Journal of Machine Learning Research}, 21(1):5485--5551.

\bibitem[{Ramirez-Orta et~al.(2022)Ramirez-Orta, Xamena, Maguitman, Milios, and Soto}]{ramirez2022post}
Juan~Antonio Ramirez-Orta, Eduardo Xamena, Ana Maguitman, Evangelos Milios, and Axel~J Soto. 2022.
\newblock Post-ocr document correction with large ensembles of character sequence-to-sequence models.
\newblock In \emph{Proceedings of the AAAI Conference on Artificial Intelligence}, volume~36, pages 11192--11199.

\bibitem[{Richter et~al.(2018)Richter, Wickes, Beser, and Marcus}]{richter2018low}
Caitlin Richter, Matthew Wickes, Deniz Beser, and Mitch Marcus. 2018.
\newblock Low-resource post processing of noisy ocr output for historical corpus digitisation.
\newblock In \emph{Proceedings of the Eleventh International Conference on Language Resources and Evaluation (LREC 2018)}.

\bibitem[{Rigaud et~al.(2019)Rigaud, Doucet, Coustaty, and Moreux}]{rigaud2019icdar}
Christophe Rigaud, Antoine Doucet, Micka{\"e}l Coustaty, and Jean-Philippe Moreux. 2019.
\newblock {ICDAR-2019} competition on post-{OCR} text correction.
\newblock In \emph{Proceedings of the 2019 IAPR International Conference on Document Analysis and Recognition (ICDAR)}, pages 1588--1593. IEEE.

\bibitem[{Rublee et~al.(2011)Rublee, Rabaud, Konolige, and Bradski}]{rublee2011orb}
Ethan Rublee, Vincent Rabaud, Kurt Konolige, and Gary Bradski. 2011.
\newblock Orb: An efficient alternative to sift or surf.
\newblock In \emph{2011 International conference on computer vision}, pages 2564--2571. Ieee.

\bibitem[{Sennrich et~al.(2015)Sennrich, Haddow, and Birch}]{sennrich2015improving}
Rico Sennrich, Barry Haddow, and Alexandra Birch. 2015.
\newblock Improving neural machine translation models with monolingual data.
\newblock \emph{arXiv preprint arXiv:1511.06709}.

\bibitem[{Shorten and Khoshgoftaar(2019)}]{shorten2019survey}
Connor Shorten and Taghi~M Khoshgoftaar. 2019.
\newblock A survey on image data augmentation for deep learning.
\newblock \emph{Journal of big data}, 6(1):1--48.

\bibitem[{Smith(2007)}]{smith2007overview}
Ray Smith. 2007.
\newblock An overview of the {Tesseract OCR} engine.
\newblock In \emph{Proceedings of the 9th International Conference on Document Analysis and Recognition (ICDAR)}, volume~2, pages 629--633. IEEE.

\bibitem[{Soper et~al.(2021)Soper, Fujimoto, and Yu}]{soper2021bart}
Elizabeth Soper, Stanley Fujimoto, and Yen-Yun Yu. 2021.
\newblock Bart for post-correction of {OCR} newspaper text.
\newblock In \emph{Proceedings of the Seventh Workshop on Noisy User-generated Text (W-NUT 2021)}, pages 284--290.

\bibitem[{Stankevi{\v{c}}ius et~al.(2022)Stankevi{\v{c}}ius, Luko{\v{s}}evi{\v{c}}ius, Kapo{\v{c}}i{\=u}t{\.e}-Dzikien{\.e}, Briedien{\.e}, and Krilavi{\v{c}}ius}]{stankevivcius2022correcting}
Lukas Stankevi{\v{c}}ius, Mantas Luko{\v{s}}evi{\v{c}}ius, Jurgita Kapo{\v{c}}i{\=u}t{\.e}-Dzikien{\.e}, Monika Briedien{\.e}, and Tomas Krilavi{\v{c}}ius. 2022.
\newblock Correcting diacritics and typos with a byt5 transformer model.
\newblock \emph{Applied Sciences}, 12(5):2636.

\bibitem[{Stubbs(1996)}]{stubbs1996text}
Michael Stubbs. 1996.
\newblock \emph{Text and corpus analysis: Computer-assisted studies of language and culture}.
\newblock Blackwell Oxford.

\bibitem[{Tang et~al.(2020)Tang, Tran, Li, Chen, Goyal, Chaudhary, Gu, and Fan}]{tang2020multilingual}
Yuqing Tang, Chau Tran, Xian Li, Peng-Jen Chen, Naman Goyal, Vishrav Chaudhary, Jiatao Gu, and Angela Fan. 2020.
\newblock Multilingual translation with extensible multilingual pretraining and finetuning.
\newblock \emph{arXiv preprint}, (2008.00401).

\bibitem[{Tarafdar et~al.(2019)Tarafdar, Beath, and Ross}]{tarafdar2019using}
Monideepa Tarafdar, Cynthia~M Beath, and Jeanne~W Ross. 2019.
\newblock Using ai to enhance business operations.
\newblock \emph{MIT Sloan Management Review}, 60(4).

\bibitem[{Vaswani et~al.(2017)Vaswani, Shazeer, Parmar, Uszkoreit, Jones, Gomez, Kaiser, and Polosukhin}]{vaswani2017attention}
Ashish Vaswani, Noam Shazeer, Niki Parmar, Jakob Uszkoreit, Llion Jones, Aidan~N Gomez, {\L}ukasz Kaiser, and Illia Polosukhin. 2017.
\newblock Attention is all you need.
\newblock \emph{Advances in neural information processing systems}, 30.

\bibitem[{Wemhoener et~al.(2013)Wemhoener, Yalniz, and Manmatha}]{wemhoener2013creating}
David Wemhoener, Ismet~Zeki Yalniz, and R~Manmatha. 2013.
\newblock Creating an improved version using noisy ocr from multiple editions.
\newblock In \emph{2013 12th International Conference on Document Analysis and Recognition}, pages 160--164. IEEE.

\bibitem[{Xue et~al.(2022)Xue, Barua, Constant, Al-Rfou, Narang, Kale, Roberts, and Raffel}]{xue2022byt5}
Linting Xue, Aditya Barua, Noah Constant, Rami Al-Rfou, Sharan Narang, Mihir Kale, Adam Roberts, and Colin Raffel. 2022.
\newblock {Byt5: Towards a token-free future with pre-trained byte-to-byte models}.
\newblock \emph{Transactions of the Association for Computational Linguistics}, 10:291--306.

\bibitem[{Xue et~al.(2020)Xue, Constant, Roberts, Kale, Al-Rfou, Siddhant, Barua, and Raffel}]{xue2020mt5}
Linting Xue, Noah Constant, Adam Roberts, Mihir Kale, Rami Al-Rfou, Aditya Siddhant, Aditya Barua, and Colin Raffel. 2020.
\newblock mt5: A massively multilingual pre-trained text-to-text transformer.
\newblock \emph{arXiv preprint arXiv:2010.11934}.

\bibitem[{Yalniz and Manmatha(2011)}]{yalniz2011fast}
Ismet~Zeki Yalniz and Raghavan Manmatha. 2011.
\newblock A fast alignment scheme for automatic {OCR} evaluation of books.
\newblock In \emph{2011 International Conference on Document Analysis and Recognition}, pages 754--758. IEEE.

\end{thebibliography}
\bibliographystyle{acl_natbib}

\appendix

\section{Training parameters}
\label{sec:appendix}
Due to the numerous sub-experiments in this study, models were trained on both RTX 4090 24GB and A100 80GB GPUs, without selecting different training parameters for different languages.

The mT5-base, ByT5-base, mBART-large, CharBERT, and ResNet50 models were all sourced from Hugging Face. Parameters for mT5-base, ByT5-base, and mBART-large were identical: dropout rate of 0.2, learning rate of 5e-4, batch size of 4, and 6 epochs. For CharBERT-related models, the dropout rate is 0.2, learning rate is 3e-5, batch size is 8, with 12 epochs. The ENSEMBLE and SCRATCH models have the same parameters: dropout rate of 0.1, learning rate of 1e-4, batch size of 32, and 30 epochs. All models use Adam optimizer and are trained in fp32 precision. The best models are selected based on dev loss.

\end{document}